\documentclass[11pt]{article}

\usepackage[preprint]{acl}

\usepackage{times}
\usepackage{latexsym}
\usepackage{booktabs}
\usepackage[table]{xcolor}
\usepackage{array}           
\usepackage{caption}
\usepackage{subcaption}
\usepackage{xurl}
\usepackage{multirow} 
\usepackage[most]{tcolorbox}
\tcbuselibrary{listings}
\usepackage{hyphenat} 

\usepackage[T1]{fontenc}

\usepackage[utf8]{inputenc}

\usepackage{microtype}

\usepackage{inconsolata}

\usepackage{graphicx}

\usepackage{amsmath}
\usepackage{amssymb}

\usepackage{subcaption}
\usepackage[most]{tcolorbox}
\usepackage{xcolor}
\usepackage[normalem]{ulem} 

\definecolor{highlightred}{HTML}{D62728}     
\definecolor{boxbg}{rgb}{0.98,0.98,0.98}     
\definecolor{boxframe}{rgb}{0.80,0.80,0.80}  
\definecolor{headerbg}{HTML}{D2D9F2}  
\definecolor{headergreen}{HTML}{B7D7B0} 
\definecolor{symbolcolor}{HTML}{0050B0} 
\definecolor{exampleheader}{HTML}{D2D9F2} 
\definecolor{failureheader}{HTML}{F2C6C2} 
\definecolor{failureheader2}{HTML}{D88C8C} 

\definecolor{mygreen}{HTML}{E8F4EC}
\definecolor{myred}{HTML}{FBEAEA} 
\definecolor{myband}{HTML}{F2F4F6}

\tcbset{
  promptbox/.style={
    colback=gray!10,
    coltitle=black,
    colbacktitle=exampleheader,
    sharp corners,
    fonttitle=\bfseries,
    boxrule=0.3pt,
    top=3pt, bottom=2pt,
    left=4pt, right=4pt,
    fontupper=\small,
    fontlower=\small,
    enhanced,
    boxsep=3pt,
    title filled
  }
}

\title{
  \fontsize{14.1}{17}\selectfont
  \textsc{EdgeLM}: Edge Demonstrations for Language Models' Table Understanding
}

\author{
\normalsize Soroush Omidvartehrani \\
\normalsize University of Alberta, Canada\\
\normalsize \texttt{s.omidvartehrani@ualberta.ca}
\And
\normalsize Mohammadamin Habibollah \\
\normalsize University of Alberta, Canada\\
\normalsize \texttt{mhabibol@ualberta.ca}
\AND
\normalsize Mohammadreza Daviran \\
\normalsize University of Alberta, Canada\\
\normalsize \texttt{daviran@ualberta.ca}
\And
\normalsize Davood Rafiei \\
\normalsize University of Alberta, Canada\\
\normalsize \texttt{drafiei@ualberta.ca}
}

\begin{document}
\maketitle
\begin{abstract}
Large language models (LLMs) perform table-centric prediction through in-context learning, making demonstration selection critical to performance. Existing retrieval methods prioritize similarity to the query, but similar demonstrations often reinforce the model's likely prediction rather than reveal the distinctions needed for difficult decisions. We propose \textsc{EdgeLM}, a retrieval framework that instead selects edge evidence, demonstrations that are both relevant to the query and informative about the decision boundary. \textsc{EdgeLM} retrieves two complementary forms of edge evidence by selecting data edges, nearby examples with different ground-truth labels, and model edges, similar examples previously misclassified by the deployed model. \textsc{EdgeLM} requires neither model retraining nor task-specific engineering. Across five data wrangling tasks, fifteen datasets, and five open-weight and proprietary LLMs, \textsc{EdgeLM} consistently achieves the best or near-best performance in every setting, while ablations show that the two forms of edge evidence provide complementary benefits.
Our code and datasets are publicly available at \url{https://github.com/soroushomidvar/EdgeLM}.
\end{abstract}

\section{Introduction}
  
Tables are widely used for analysis, decision making, and machine learning, but real-world tabular data rarely comes ready for use. Values may be missing, records that refer to the same entity may not match, and related columns across tables may use different names or formats. Before the data can be used reliably, practitioners must detect erroneous cells, identify duplicate entities, align schemas, fill missing values, and flag anomalous records. These operations, collectively known as \emph{data wrangling}, are fundamental to ensuring data quality and directly affect the reliability of downstream analytics, machine learning models, and AI systems.

Automating data wrangling remains challenging because different tasks require different forms of reasoning over heterogeneous tables. Determining whether a cell is erroneous, whether two records refer to the same entity, or whether two attributes should be aligned often requires reasoning over relationships among cells, rows, and columns rather than individual values in isolation. Consequently, existing systems typically rely on task-specific rules, handcrafted features, weak supervision, or separately trained models~\cite{mahdavi2019raha,mahdavi2020baran,li2020deep}. Although effective in specialized settings, these approaches require substantial engineering effort and often generalize poorly across new tasks, schemas, and domains.

LLMs offer a promising alternative by leveraging pretrained knowledge to tackle diverse data wrangling tasks with a single model~\cite{narayan2022can,li2024table,zhang2024jellyfish}. However, their performance depends critically on the demonstrations provided during in-context learning or the examples seen during fine-tuning~\cite{min2022rethinking, zhao2021calibrate,
     liu2022makes}. This raises a fundamental question: \textit{what makes a demonstration effective across diverse table-centric prediction tasks?}

Selecting demonstrations is far from straightforward. Including many rows can exceed the context budget while diluting relevant evidence, whereas manually curated examples do not scale to unseen tables and queries. Random sampling and diversity-based selection often ignore the query~\cite{su2023selective}, while similarity-based retrieval favors examples that resemble the query~\cite{liu2022makes,rubin2022learning} but often fails to expose the evidence that distinguishes competing prediction. For example, two entity pairs may appear nearly identical except for a subtle mismatch that changes the label, or a candidate cell value may look plausible yet violate a dependency elsewhere in the row. If all retrieved demonstrations support the same prediction, they reinforce the likely answer without revealing where the decision boundary lies. Effective demonstrations should therefore be not only relevant, but also \emph{discriminative of the local decision boundary}.

We hypothesize that demonstrations near decision boundaries provide the missing discriminative signal. Similar examples establish the local context, whereas boundary examples reveal the evidence that distinguishes competing predictions. Motivated by this observation, we propose \textsc{EdgeLM}, a task-agnostic framework for selecting demonstrations for structured prediction over tabular data. EdgeLM augments similarity-based retrieval with \emph{edge evidence}: demonstrations that are both relevant to the query and informative about the local decision boundary. We instantiate edge evidence through two complementary perspectives: the \emph{data edge}, which retrieves neighboring examples from different gold labels, and the \emph{model edge}, which retrieves nearby examples that the deployed LLM predicts incorrectly, using them as corrective evidence. The selected demonstrations fit within a fixed prompt budget and require neither model retraining nor task-specific engineering.

We evaluate EdgeLM on five representative data wrangling tasks, fifteen datasets, and five open-weight and proprietary LLMs. Across nearly all task--model combinations, EdgeLM achieves the best performance or performs within one standard deviation of the best result. Ablation studies further show that the data edge and model edge provide complementary benefits across different allocations of the demonstration budget.

  \section{Related Work}

  \subsection{LLMs for Table Understanding and Data Wrangling}
  
  Large language models have substantially broadened the range of tabular reasoning tasks that can be addressed with a single underlying model. Early work adapted language models to tables through table-aware pretraining~\citep{yin2020tabert, herzig2020tapas, liu2022tapex, borisov2023great}, while more recent approaches leverage instruction tuning and general-purpose LLMs~\citep{li2024table, zhang2024tablellama, zhang2024jellyfish}. Other methods improve table understanding through multi-step reasoning and retrieval, such as Chain-of-Table~\citep{wang2024chainoftable} and TableRAG~\citep{chen2024tablerag}. Meanwhile, studies of serialization and row--column ordering have shown that LLM performance is highly sensitive to the context presented to the prompt~\citep{sui2024table, liu2024rethinking, habibollah2026synsql, yang2022tableformer, fang2024survey}.

  
  LLMs have also been applied to individual data wrangling tasks, including data cleaning~\citep{naeem2024retclean}, entity resolution~\citep{peeters2023chatgpt, peeters2024entity}, anomaly detection~\citep{li2024anomaly}, and missing-value imputation~\citep{wang2025llm}. Across these and related tabular and
database problems, existing approaches commonly rely on task or
schema-specific prompts, retrieval strategies, or adaptation
techniques~\citep{nan-etal-2023-enhancing, he-etal-2025-llm, daviran2026sqlexchange}. In contrast, we study demonstration selection itself as a task-agnostic problem, using a unified formulation that spans multiple data wrangling tasks while keeping the model, prompt structure, and inference procedure fixed.


  \subsection{Demonstration Selection for In-Context Learning}
%
  The choice of demonstrations has a substantial impact on in-context learning~\citep{min2022rethinking, zhao2021calibrate}.  Existing methods primarily optimize demonstration selection for relevance, coverage, or uncertainty. Similarity-based retrieval selects examples that resemble the input query~\citep{liu2022makes, rubin2022learning}, whereas diversity-based methods seek broader coverage of the input or output space~\citep{su2023selective, levy2023diverse}. Other approaches combine relevance and diversity, iteratively refine the selected set, or explicitly balance label distributions~\citep{qin2024iterative, omidvartehrani2026ldilocalizeddataimputation, tran-etal-2025-improving}. Uncertainty-guided methods instead identify globally informative examples for annotation or prompting~\citep{diao-etal-2024-active, mavromatis-etal-2024-covericl}. 

  Our work addresses a complementary question: \emph{which nearby demonstrations are most informative for distinguishing competing predictions?} Existing retrieval methods favor examples that are representative of the query but do not explicitly retrieve demonstrations that lie near local decision boundaries or reveal systematic errors made by the deployed model. EdgeLM fills this gap by retrieving demonstrations that are both relevant and discriminative through complementary \emph{data-edge} and \emph{model-edge} evidence.

\section{Preliminary} \label{sec:preliminary}
Before presenting \textsc{EdgeLM}, we introduce a unified formulation for prediction over structured tabular data that abstracts away individual tasks. Although our evaluation focuses on data wrangling tasks, the formulation naturally extends to a broader class of table-centric prediction problems.

\smallskip
\noindent\textbf{Unified Prediction Formulation.}
We consider prediction problems defined over one or more relational tables. Given a query instance extracted from the tables, the objective is to predict its unknown label. This abstraction encompasses a broad range of table-centric prediction problems, including data wrangling, table understanding, and related structured prediction tasks.

Formally, let $\tau$ denote a prediction task with instance space $\mathcal{X}_\tau$ and finite label space $\mathcal{Y}_\tau$. Given a query instance $x\in\mathcal{X}_\tau$, the goal is to predict
\[
y=f_\tau^*(x)\in\mathcal{Y}_\tau,
\]
where $f_\tau^*:\mathcal{X}_\tau\rightarrow\mathcal{Y}_\tau$ is the unknown task-specific decision function.

The form of the query instance depends on the application. It may correspond to a cell, a tuple, a pair of tuples, a pair of attributes, an entire table, or any other structured object derived from one or more relational tables. Let $\mathcal{D}$ denote a table with schema $\mathcal{R}=\{A_1,\ldots,A_m\}$, where each tuple $t_i\in\mathcal{D}$ contains attribute values $t_i[A_j]$. Tasks involving multiple tables additionally assume a second table $\mathcal{D}'$ with schema $\mathcal{R}'$.

Table~\ref{tab:tasks} illustrates five representative data wrangling tasks that instantiate this formulation. These tasks serve as evaluation benchmarks throughout the paper, but the formulation itself is considerably more general.

\begin{table}[tb]
\centering
\small
\setlength{\tabcolsep}{3pt}
\caption{Representative table-centric prediction tasks under the unified prediction formulation.}
\label{tab:tasks}
\begin{tabular}{@{}llll@{}}
\toprule
Task & Instance $x$ & Labels $\mathcal{Y}_\tau$ & Question ($f^\ast_\tau$)$^{\dagger}$ \\
\midrule
ED & $(t_i[A_j],\, t_i)$ & $\{\textsf{error},\textsf{clean}\}$     & Cell corrupted? \\
EM & $(t,\, t')$         & $\{\textsf{match},\textsf{non-match}\}$ & Same entity? \\
SM & $(A,\, A')$         & $\{\textsf{match},\textsf{non-match}\}$ & Same attribute? \\
DI & $(t_i,\, A_T)$      & $\mathrm{dom}(A_T)$                     & Which value? \\
AD & $t_i$               & $\{\textsf{normal},\textsf{anomaly}\}$  & Anomalous? \\
\bottomrule
\end{tabular}

\vspace{2pt}
{\footnotesize\raggedright $^{\dagger}$The exact statement of each is given in Appendix~\ref{app:prompts}.\par}
\vspace{-4mm}
\end{table}

\smallskip
\noindent \textbf{Retrieval-Augmented Prediction.}
Rather than adapting the language model to each task, we perform prediction through in-context learning using a frozen LLM $\mathcal{M}$. The central design choice is therefore not the language model itself, but which labeled instances are retrieved and rendered as demonstrations. For each task, we assume access to a labeled reference pool
\[
\mathcal{P}_\tau\subseteq\mathcal{X}_\tau\times\mathcal{Y}_\tau,
\]
whose elements are labeled instances drawn from the same application domain as the query. Depending on the application, they may correspond to labeled cells, tuples, tuple pairs, attribute pairs, completed tuples, or other structured objects.           

Given a query instance $x$, a retrieval function
$\sigma(x,\mathcal{P}_\tau)$
selects a set of $k$ instances
$\mathcal{S}\subseteq\mathcal{P}_\tau.$
Each retrieved instance is rendered as an in-context demonstration and included together with $x$ in the prompt supplied to $\mathcal{M}$. The model predicts
\[
\hat{y}
=
\arg\max_{y\in\mathcal{Y}_\tau}
p_{\mathcal{M}}(y\mid x,\mathcal{S}),
\]
and the prediction is considered correct when $\hat{y}=f_\tau^*(x)$.


\smallskip
\noindent \textbf{Goal.}
Rather than modifying the underlying language model $\mathcal M$, we focus on selecting the demonstration set $\mathcal S$.  Our objective is to improve prediction accuracy solely through demonstration selection, while keeping the language model, prompts, and task formulation fixed. Specifically, we seek a retrieval function $\sigma$ that selects demonstrations that are both relevant to the query and informative about the local decision boundary of the unknown decision function $f_\tau^*$. Although our experiments focus on data wrangling, the proposed retrieval strategy applies to any table-centric prediction problem that can be solved through in-context learning from labeled examples.



\section{Edge-Aware Demonstration Selection}
\label{sec:method}

Section~\ref{sec:preliminary} formulated demonstration selection as the problem of retrieving examples that are both relevant to the query and informative about the local decision boundary of the unknown decision function $f_\tau^*$. Because this boundary is not directly observable, \textsc{EdgeLM} approximates it using two complementary signals: label variation in the local data neighborhood and errors made by the deployed language model. Similarity-based retrieval preserves relevance to the query, while these two forms of \emph{edge evidence} expose distinctions that conventional nearest-neighbor retrieval may overlook.

\begin{figure}[tb]
    \centering
    \captionsetup{font=footnotesize}
    \begin{subfigure}[b]{0.47\columnwidth}
        \centering
        \includegraphics[width=\textwidth]{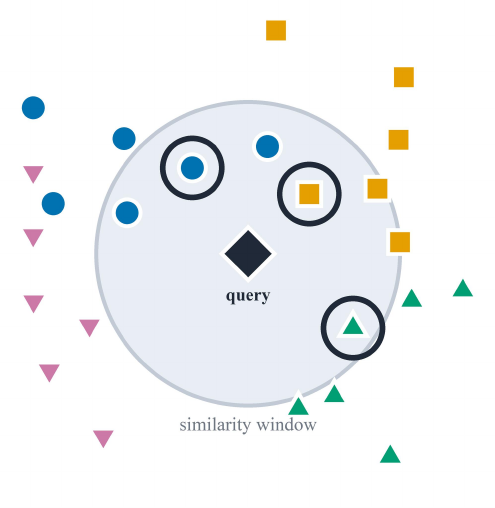}
        \caption{Similar examples, different labels}
        \label{fig:edge_views_a}
    \end{subfigure}
    \hfill
    \raisebox{1.2cm}{\rule{0.4pt}{2.5cm}}
    \hfill
    \begin{subfigure}[b]{0.47\columnwidth}
        \centering
        \includegraphics[width=\textwidth]{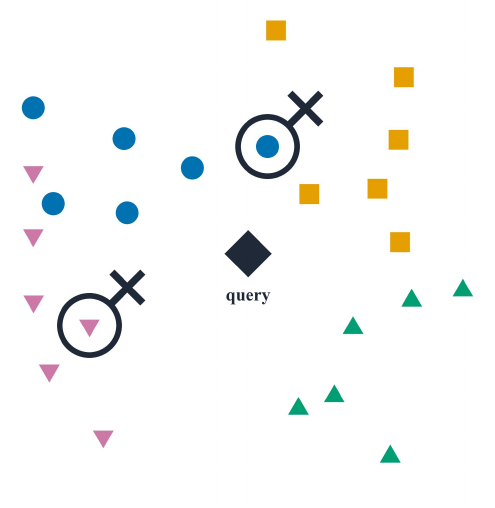}
        \caption{Close examples model got wrong}
        \label{fig:edge_views_b}
    \end{subfigure}
    \caption{Two complementary sources of edge evidence in \textsc{EdgeLM}: (a) nearby examples with different labels, and (b) nearby examples that the LLM previously predicted incorrectly.}
    \label{fig:edge_view}
\vspace{-4mm}
\end{figure}

The first signal comes from the labeled data. Nearby instances with different ground-truth labels indicate that competing outcomes occur within the same local region of the instance space. We call demonstrations selected from such local label contrasts \emph{data-edge demonstrations}. The second signal comes from the deployed model. Nearby instances that the model predicts incorrectly identify regions where its approximation of $f_\tau^*$ diverges from the ground truth. We call these \emph{model-edge demonstrations}. Thus, data edges characterize local variation in the task labels, whereas model edges characterize the model's local failure modes (see Figure~\ref{fig:edge_view}).



\begin{figure*}[t]

    \centering

    \includegraphics[width=\textwidth]{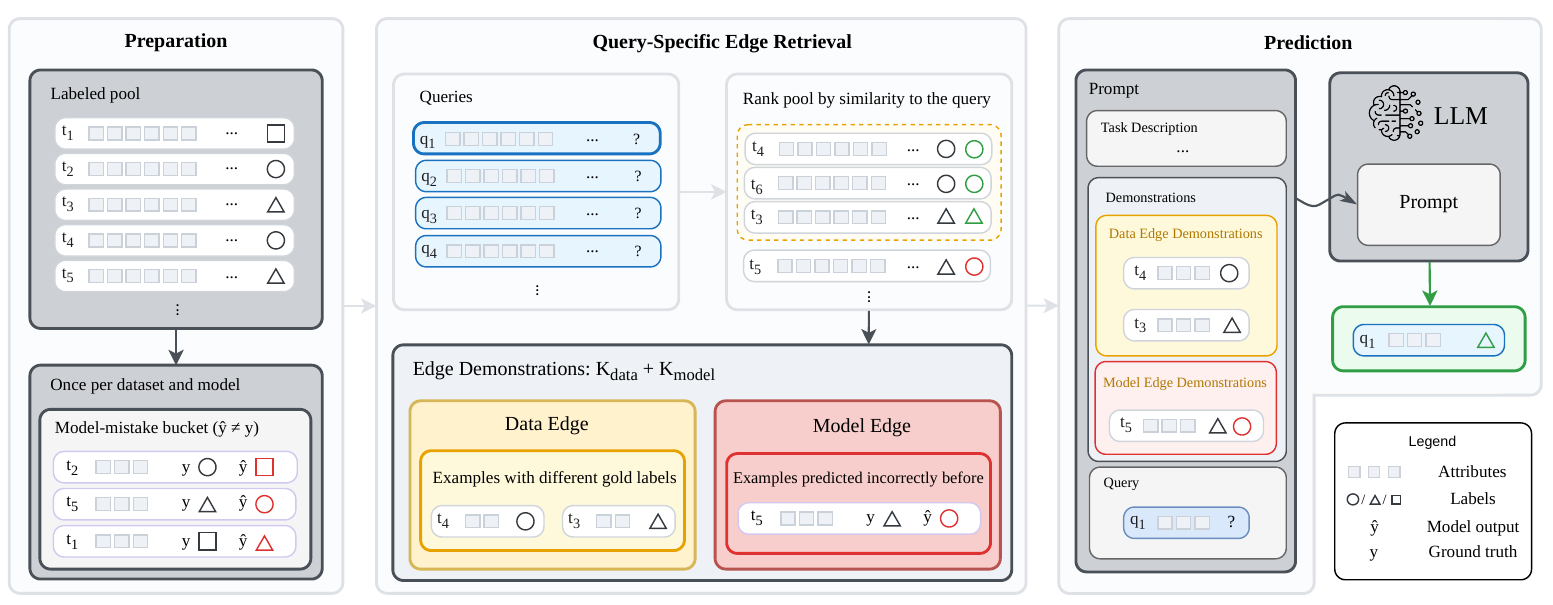}

    \caption{Overview of \textsc{EdgeLM}. Before inference, the LLM is evaluated zero-shot on the labeled pool to identify model-specific errors. For each test query, pool instances are ranked by similarity. \textsc{EdgeLM} selects locally contrasting examples from the query neighborhood (data edge) and retrieves similar model errors from the precomputed error pool (model edge). The two sets are combined into a single prompt and passed to the LLM.}

    \label{fig:EdgeLM-overview}
\vspace{-4mm}
\end{figure*}

Both components use the same task-specific similarity function as conventional retrieval and differ only in how they select from the ranked candidates.
Given a total demonstration budget $K$, \textsc{EdgeLM} allocates $K_{data}$ positions to local label contrast and $K_{model}$ positions to model-specific errors, with $K_{data} + K_{model} = K$. The two sets are combined into a single prompt,  as summarized in Figure~\ref{fig:EdgeLM-overview}. The following subsections describe how each source of edge evidence is constructed and how the resulting demonstrations are assembled.








\subsection{Data Edge: Local Label Contrast}

\label{sec:local-contrast}

The data-edge component seeks demonstrations that introduce label contrast without sacrificing relevance. 
For a query $x$, we rank the labeled pool $\mathcal{P}_\tau$ by similarity to $x$ and define  $\mathcal{N}_w(x)$ as the neighborhood containing the top $w$ ranked instances. Restricting selection to $\mathcal{N}_w(x)$ ensures that any introduced contrast remains local to the query.

A conventional similarity-based strategy would select the $K_{\mathrm{data}}$ nearest instances. This set may, however, be label-homogeneous even when other labels occur slightly farther away within the same neighborhood. For instance, the nearest entity pairs may all have the label \textit{match}, repeatedly reinforcing that prediction without showing which subtle differences would instead imply \textit{non-match}. Such demonstrations establish similarity but provide little evidence about the distinction between competing outcomes.

\textsc{EdgeLM} intervenes only when the nearest-neighbor set fails to represent the label contrast available locally. We first inspect the $K_{\mathrm{data}}$ nearest instances. If they already cover the maximum number of distinct labels that can be represented under the available budget and among the labels observed in $\mathcal{N}_W(x)$, they are retained unchanged. Otherwise, the instances in $\mathcal{N}_W(x)$ are partitioned by their ground-truth labels, with instances within each label group ordered by similarity to $x$. The groups are ordered by the similarity of their nearest member and visited in round-robin order, selecting the most similar remaining instance from each group until $K_{\mathrm{data}}$ instances have been chosen.


We denote the resulting sequence by $\mathcal{S}_{\mathrm{data}}(x)$. This procedure applies to both binary and multi-class tasks. Importantly, its objective is not to balance the labels globally, but to expose label contrast within the query's local neighborhood. If $\mathcal{N}_W(x)$ contains only one label, no local contrast is available and the procedure reduces to ordinary nearest-neighbor selection.

\subsection{Model Edge: Query-Local Error Retrieval}

\label{sec:model-errors}

The model-edge component provides a complementary, model-specific view of the boundary. Even when the labeled neighborhood contains little label contrast, nearby examples on which the deployed model fails can reveal distinctions that the model has not learned to make reliably ~\citep{kadavath2022language}. 
Because different language models may fail on different instances of the same task, this evidence captures local weaknesses that cannot be obtained from the labeled data alone.



Before test-time inference, we evaluate the model zero-shot on every labeled
instance $(x_i,y_i) \in \mathcal{P}_\tau$. Let
$\hat{y}_i^{\mathrm{zs}}$ denote the prediction obtained
using the same task instruction and inference procedure as at test time, 
but without demonstrations: $\mathcal{M}\!\left(x_i,\varnothing\right)$.
We then construct the model-specific error pool:
\[
\mathcal{E}_{\mathcal{M},\tau}
=
\left\{
\left(x_i,\hat{y}_i^{\mathrm{zs}},y_i\right)
\;:\;
(x_i,y_i)\in\mathcal{P}_\tau,
\;
\hat{y}_i^{\mathrm{zs}}\neq y_i
\right\}
\].
The error pool is constructed once for each model, task, and dataset using only the labeled reference pool; test instances are never involved.\footnote{Appendix~\ref{app:cost} quantifies the cost of this step and shows that the added per-query retrieval stays within a small factor of plain similarity retrieval and well under a millisecond.} 

For a query $x$, we rank the instances in $\mathcal{E}_{\mathcal{M},\tau}$ by their similarity to $x$ and select the $K_{model}$ nearest errors. The resulting sequence, denoted by $\mathcal{S}_{\mathrm{model}}(x)$, contains model failures that are locally relevant to the query rather than arbitrary errors from elsewhere in the instance space.

Each selected error is rendered as a corrective demonstration containing the original instance, the model's incorrect zero-shot prediction, and the corresponding ground-truth label. This format explicitly exposes the distinction that the model previously failed to make. These examples should not be interpreted as known boundary points of the underlying function $f_\tau^*$; rather, they identify local regions where the deployed model disagrees with the available supervision. The complete prompt format is provided in Appendix~\ref{app:prompts}.

\subsection{Edge Prompt Construction}
\label{sec:edge-assembly}


For each query $x$,  \textsc{EdgeLM} concatenates the data-edge and model-edge sequences:
$
\mathcal{S}_{\mathrm{edge}}(x)
=
\mathcal{S}_{\mathrm{data}}(x)
\mathbin{\|}
\mathcal{S}_{\mathrm{model}}(x),
$
where $\mathbin{\|}$ denotes sequence concatenation.  Data-edge demonstrations precede model-edge demonstrations in all experiments. To avoid redundancy, instances already selected for $\mathcal{S}_{\mathrm{data}}(x)$ are excluded when constructing $\mathcal{S}_{\mathrm{model}}(x)$, ensuring that no instance occupies multiple prompt positions. 
The total demonstration budget $K$ is divided according to a fixed $K_{data}:K_{model}$ ratio, whose sensitivity is examined in Section~\ref{sec:robustness}.

Instances in $\mathcal{S}_{\mathrm{data}}(x)$ are rendered as standard input--answer demonstrations. Each instance in $\mathcal{S}_{\mathrm{model}}(x)$ additionally includes the model's incorrect zero-shot prediction before presenting the ground-truth answer. The final prompt consists of the task instruction,  the combined edge demonstrations, and the query instance, in that order. The prompt is then supplied to the LLM to obtain the prediction.

\section{Experiments} \label{sec:experiments}
This section, together with the supporting appendices, presents a comprehensive evaluation of our proposed demonstration selection framework. The experiments aim to answer four main questions: (1)~Without any fine-tuning, does drawing demonstrations from the decision boundary beat zero-shot, random, label-diverse, and similarity selection, does that advantage hold across tasks, datasets, model families and scales, and where does it weaken (\S\ref{sec:main-results}, Appendix~\ref{app:perdataset})? (2)~Does each of the two edge views help on its own, or is one of them carrying the gain (\S\ref{sec:ablation})? (3)~How much does the gain depend on how the demonstration budget is divided between the two views (\S\ref{sec:robustness})? (4)~Is the method robust to hyperparameters, the budget $K$ and the selection window $w$, and what does it cost to run (Appendices~\ref{app:cost} and \ref{app:kw})?

\subsection{Datasets and Experimental Setup}
Table~\ref{tab:datasets} summarizes the datasets. They vary in domain, size, and number of attributes, and are widely used in data wrangling studies~\cite{zhang2024jellyfish, mahdavi2019raha, li2020deep, narayan2022can, kopcke2010evaluation, koutras2021valentine, vanschoren2014openml, mahdavi2020baran}, providing a diverse testbed. Each dataset is subsampled to at most $2{,}500$ instances and split $80/20$ into a demonstration pool and a test set; the pool is the only source of demonstrations, and both subsample and split are fixed by a common random
seed, so all methods and models are evaluated on identical test instances. We evaluate five LLMs of varying sizes from five families: three open-weight (Qwen2.5-7B, Llama-3.1-8B, Mistral-Nemo-12B) and two proprietary (GPT-4o-mini, Gemini-3.1-flash-lite). On every model we compare five demonstration-selection strategies: \emph{zero-shot} (no demonstrations), \emph{random} (drawn uniformly from the pool), \emph{diverse} (drawn to balance the label space), \emph{similarity} (the query's nearest neighbors), and \textsc{EdgeLM} (ours). Prior task-specific systems couple the choice of demonstrations with other design choices, such as a tailored prompt, a specialised model, or per-task tuning, and are typically developed for one or a few tasks; the demonstration strategy is therefore one ingredient among several, and its individual effect is hard to read off directly. Because our aim is to measure the effect of demonstration selection on its own, we hold everything else fixed: within each model the five strategies use an identical prompt template, label space, demonstration pool, and test set, and differ only in which demonstrations occupy the $K$ slots. Unless otherwise specified, each uses $K\!=\!5$ demonstrations, and we report accuracy and F1 (weighted F1 for the multi-class DI). Each configuration is averaged over three seeds, the open-weight models run on a single H100 GPU on an HPC cluster, and the proprietary models are queried through their providers' APIs.

\begin{table}[t]
\centering
\small
\setlength{\tabcolsep}{3pt}
\caption{The 15 datasets. \#inst is the task's instance (cell for ED, record pair for EM, column pair for SM, row for DI, record for AD).}
\label{tab:datasets}
\begin{tabular}{@{}lllrr@{}}
\toprule
Task & Dataset & Domain & \#attr & \#inst \\
\midrule
\multirow{3}{*}{ED} & Beers          & beer catalog     & 11 & 26{,}510 \\
                    & Flights        & flight times     & 7  & 16{,}632 \\
                    & Hospital       & hospital quality & 20 & 20{,}000 \\
\midrule
\multirow{3}{*}{EM} & Amazon-Google  & software         & 3  & 11{,}460 \\
                    & DBLP-ACM       & bibliographic    & 4  & 12{,}363 \\
                    & Walmart-Amazon & retail products  & 5  & 10{,}242 \\
\midrule
\multirow{3}{*}{SM} & ChEMBL-Join    & chemical assays  & 2  & 1{,}200 \\
                    & OpenData-Join  & open data        & 2  & 1{,}200 \\
                    & TPC-DI-Join    & finance          & 2  & 1{,}188 \\
\midrule
\multirow{3}{*}{DI} & Splice         & DNA sequences    & 60 & 3{,}190 \\
                    & DiabetesPima   & clinical         & 8  & 768 \\
                    & BalanceScale   & balance scale    & 4  & 625 \\
\midrule
\multirow{3}{*}{AD} & KDD-DoS        & DoS attacks      & 41 & 391{,}458 \\
                    & KDD-Probe      & probing attacks  & 41 & 4{,}107 \\
                    & KDD-RareAttack & rare attacks     & 41 & 1{,}178 \\
\bottomrule
\end{tabular}
\vspace{-4mm}
\end{table}

\begin{table}[t]
\centering
\caption{Demonstration selection across different models. Each cell is F1 / weighted F1 (Accuracy in parentheses), and \textbf{bold} marks the best F1 per row and any within seed-to-seed standard deviation of it, i.e.\ a tie.}
\label{tab:main}
\setlength{\tabcolsep}{2pt}
\resizebox{\columnwidth}{!}{\footnotesize
\begin{tabular}{@{}lccccc@{}}
\toprule
Task & Zero-shot & Random & Diversity & Similarity & Edge \\
\midrule
\multicolumn{6}{@{}l}{\textit{Qwen2.5-7B}} \\
\midrule
ED & .322 (.813) & .237 (.778) & .251 (.795) & .279 (.784) & \textbf{.388 (.819)} \\
EM & .701 (.941) & .643 (.907) & .658 (.908) & .738 (.945) & \textbf{.772 (.955)} \\
SM & .796 (.886) & .807 (.886) & .820 (.891) & \textbf{.888 (.926)} & \textbf{.896 (.935)} \\
DI & .326 (.414) & .573 (.568) & .543 (.540) & \textbf{.588 (.591)} & \textbf{.583 (.578)} \\
AD & .141 (.720) & .557 (.734) & .593 (.759) & .628 (.793) & \textbf{.840 (.924)} \\
\midrule
\multicolumn{6}{@{}l}{\textit{Llama-3.1-8B}} \\
\midrule
ED & .286 (.669) & .345 (.816) & .412 (.822) & .474 (.771) & \textbf{.581 (.849)} \\
EM & .457 (.668) & .463 (.729) & .483 (.709) & .558 (.855) & \textbf{.627 (.900)} \\
SM & .789 (.875) & .800 (.866) & .799 (.869) & .910 (.936) & \textbf{.936 (.955)} \\
DI & .090 (.210) & .482 (.495) & .468 (.482) & \textbf{.586 (.594)} & \textbf{.591 (.603)} \\
AD & .346 (.741) & .551 (.742) & .575 (.697) & .810 (.906) & \textbf{.980 (.990)} \\
\midrule
\multicolumn{6}{@{}l}{\textit{Mistral-Nemo-12B}} \\
\midrule
ED & .236 (.857) & .321 (.798) & .349 (.814) & .344 (.756) & \textbf{.471 (.818)} \\
EM & .660 (.906) & .591 (.859) & .605 (.863) & .651 (.898) & \textbf{.670 (.911)} \\
SM & .786 (.870) & .823 (.872) & .854 (.901) & .787 (.836) & \textbf{.872 (.907)} \\
DI & .409 (.551) & \textbf{.524 (.539)} & .493 (.501) & .502 (.501) & .502 (.501) \\
AD & .369 (.780) & .549 (.641) & .576 (.668) & .756 (.848) & \textbf{.855 (.916)} \\
\midrule
\multicolumn{6}{@{}l}{\textit{GPT-4o-mini}} \\
\midrule
ED & .391 (.775) & .341 (.803) & .363 (.814) & .529 (.821) & \textbf{.544 (.881)} \\
EM & .767 (.955) & .693 (.915) & .711 (.923) & \textbf{.793 (.955)} & \textbf{.788 (.957)} \\
SM & .850 (.907) & .843 (.904) & .830 (.896) & .913 (.941) & \textbf{.948 (.964)} \\
DI & .226 (.283) & .553 (.540) & .545 (.537) & .555 (.556) & \textbf{.609 (.607)} \\
AD & .427 (.792) & .645 (.817) & .654 (.820) & .889 (.941) & \textbf{.955 (.978)} \\
\midrule
\multicolumn{6}{@{}l}{\textit{Gemini-3.1-flash-lite}} \\
\midrule
ED & .437 (.908) & .478 (.906) & .732 (.927) & .774 (.960) & \textbf{.834 (.963)} \\
EM & .781 (.951) & .795 (.952) & .806 (.957) & .843 (.968) & \textbf{.871 (.976)} \\
SM & .921 (.946) & .954 (.967) & .938 (.955) & \textbf{.978 (.985)} & \textbf{.984 (.989)} \\
DI & .645 (.704) & .734 (.749) & .728 (.728) & .770 (.771) & \textbf{.799 (.802)} \\
AD & .699 (.863) & .768 (.888) & .803 (.898) & \textbf{.983 (.992)} & \textbf{.991 (.996)} \\
\bottomrule
\end{tabular}}
\vspace{-4mm}
\end{table}

\subsection{Consistency Across Models and Tasks}
\label{sec:main-results}


Table~\ref{tab:main} reports the performance of five models across different demonstration-selection strategies. Across all models and tasks, \textsc{EdgeLM} achieves the highest F1 or performs comparably to the best method in nearly every setting. Its boundary-aware demonstrations are at least as effective as the best alternative on every task and every model, and usually strictly better, whereas the choice of which baseline comes closest changes from task to task and model to model.

The four baselines fall in a consistent order. Zero-shot is the weakest on average and the most volatile. It can be respectable when the task matches the model's pretraining (GPT-4o-mini reaches $0.850$ F1 on SM) yet collapse on the next task (the same model scores only $0.226$ on DI). Random improves on zero-shot by supplying the task format but adds no targeted signal. Diversity balances the label space and adds one piece of targeted structure, guaranteeing that every label appears, lifting the rare-class F1 above random on the imbalanced tasks. For example on anomaly detection, diversity beats random on all three open models (Qwen2.5-7B $0.593$ vs.\ $0.557$, Llama-3.1-8B $0.575$ vs.\ $0.551$, Mistral-Nemo-12B $0.576$ vs.\ $0.549$). Yet without locality it still trails similarity on the tasks whose decision is local, and it never approaches edge. Similarity is the strongest baseline, as expected, because near-neighbors share surface features with the query, and it leads some schema-matching and anomaly-detection cells. Nevertheless, similarity is systematically beaten on the tasks whose decision is not local; on error detection, \textsc{EdgeLM} improves F1 by $0.107$--$0.127$ across the three open models, while on entity matching the improvement reaches $0.069$. \textsc{EdgeLM} keeps similarity's locality where it helps and adds boundary and mistake demonstrations where it does not. It exceeds the diversity baseline in all $25$ model--task combinations, indicating that label diversity alone, without locality and model-specific corrections, does not close the gap.

Across tasks, the size of the gain tracks the headroom the strongest baseline leaves. Anomaly detection, whose minority (anomaly) class is rare and easily missed,
shows the largest improvements, where edge lifts F1 over similarity by $+0.212$ (Qwen2.5-7B), $+0.170$ (Llama-3.1-8B) and $+0.099$ (Mistral-Nemo-12B), and over zero-shot by as much as $+0.699$. Error detection behaves the same way, with edge ahead of similarity by roughly $0.11$ to $0.13$ F1 on the open models. At the other extreme, schema matching is close to saturated, similarity already clearing $0.91$ F1 on Llama-3.1-8B, GPT-4o-mini and Gemini-3.1-flash-lite, so edge's gains there are small ($+0.006$ to $+0.085$) and occasionally a tie, but never a loss. These improvements are concentrated in F1 rather than accuracy. For instance, on Qwen2.5-7B anomaly detection, edge raises F1 from $0.141$ to $0.840$ while accuracy moves only from $0.720$ to $0.924$, confirming that the method recovers the minority class that imbalance-blind accuracy hides.

Grouping by model type sharpens the scaling picture. On the open 7 to 12B models, whose baselines leave ample room, edge delivers its largest margins and wins or ties all five tasks on Llama-3.1-8B. On the two proprietary models the baselines are stronger, especially similarity (Gemini-3.1-flash-lite already reaches $0.978$ on SM and $0.983$ on AD), so the absolute margins compress; yet edge still finishes best or tied on all five tasks for
Gemini-3.1-flash-lite and on four of five for GPT-4o-mini, including a clear DI win ($+0.054$ F1) that none of the open models achieve. The benefit therefore does not depend on a weak model, as the baseline improves, edge's advantage narrows but does not reverse.

Finally, the table exposes why a single baseline cannot be trusted in practice. Each baseline has a task at which it looks competitive, but none carries that strength across the suite. Zero-shot is close to the top on GPT-4o-mini entity matching yet worst by a wide margin on that model's data imputation; random posts the best data-imputation score on Mistral-Nemo-12B yet sits $0.306$ F1 below edge on the same model's anomaly detection; the label-diverse baseline climbs above similarity on Mistral-Nemo schema matching ($0.854$ versus $0.787$) yet is mid-pack elsewhere; similarity dominates schema matching and anomaly detection yet is overtaken on error detection and entity matching. A practitioner committing to any one rule would win a task and lose others, and the winning choice shifts from model to model. \textsc{EdgeLM} is the only strategy at or above the best baseline on every task of every model, removing the need to guess which selection rule fits a given model--task pair. 


\subsection{Ablation Study}
\label{sec:ablation}

We isolate the two edge components by letting each one alone fill the
entire budget and comparing it with the same randomly drawn demonstrations
(Table~\ref{tab:ablation}). This construction is preferable to leave-one-out for two reasons. 
First, assigning the full budget to each component reveals its standalone effect. In a complete edge prompt, a component may occupy only a few slots or contribute little, or even nothing, for a particular dataset or model. Removing it would therefore produce only a small change and could understate its value. Second, we compare each component with $K$ random demonstrations rather than with the full edge prompt after removing that component's examples, ensuring that both prompts contain exactly $K$ demonstrations. Otherwise, the prompts would differ in length, confounding the \emph{content} of the examples with their \emph{number} and making it possible to attribute a gain to quantity rather than to the examples themselves.

Table~\ref{tab:ablation} shows that both components are effective in isolation, with each improving over the random baseline on all five tasks. The improvements range from $+0.095$ to $+0.259$ F1 for the data edge and from $+0.049$ to $+0.300$ for the model edge. Their relative strength depends on the task; the data edge contributes most to schema matching and data imputation, where a contrasting neighbor sharpens the decision boundary, whereas the model edge contributes most to error and anomaly detection, where correcting the model's own mistakes is most informative. As neither component is redundant, \textsc{EdgeLM} combines them.

\begin{table}[tb!]
\centering
\caption{Ablation. F1 / Weighted F1 for each component alone versus $K$ random, with the gain over random in parentheses.}
\label{tab:ablation}
\small
\begin{tabular*}{\columnwidth}{@{\extracolsep{\fill}}lccc@{}}
\toprule
 & Random & \multicolumn{2}{c}{Edge component alone} \\
\cmidrule(lr){3-4}
Task & F1 & Data edge & Model edge \\
\midrule
ED & 0.345 & \cellcolor{mygreen}0.474 (+0.129) & \cellcolor{mygreen}0.543 (+0.198) \\
EM & 0.463 & \cellcolor{mygreen}0.558 (+0.095) & \cellcolor{mygreen}0.572 (+0.109) \\
SM & 0.800 & \cellcolor{mygreen}0.910 (+0.110) & \cellcolor{mygreen}0.856 (+0.056) \\
DI & 0.482 & \cellcolor{mygreen}0.587 (+0.105) & \cellcolor{mygreen}0.531 (+0.049) \\
AD & 0.551 & \cellcolor{mygreen}0.810 (+0.259) & \cellcolor{mygreen}0.851 (+0.300) \\
\bottomrule
\end{tabular*}
\vspace{-4mm}
\end{table}

\subsection{Hyperparameter Robustness}
\label{sec:robustness}

\textsc{EdgeLM} exposes one substantive hyperparameter: how the $K$-demonstration budget is divided between the data-edge and the model-edge components. Table~\ref{tab:allocation} sweeps every split of the $K\!=\!5$ budget on each dataset, from all data edge ($5/0$) to all model edge ($0/5$), and compares the best split per dataset with a single fixed $2/3$ default.

The method is insensitive to this choice. The oracle-best split, selected per dataset, improves over the fixed $2/3$ default by only $+0.010$ F1 on average and by at most $+0.043$, and on five of the fifteen datasets the $2/3$ default is already optimal. The interior splits ($4/1$, $3/2$, $2/3$) lie within a few thousandths of one another, and F1 degrades only at the extremes, where one component is removed entirely ($5/0$ or $0/5$). This mirrors the ablation of Section~\ref{sec:ablation}, where both components contribute, so discarding either one costs accuracy, but the exact ratio between them does not.

This robustness is a practical advantage. Because F1 is flat across the interior allocations, applying \textsc{EdgeLM} to a new dataset or model needs no grid search over the split and no labeled data beyond the demonstration pool it already relies on. A single fixed default recovers nearly all of the accuracy. The gains in Table~\ref{tab:main} therefore do not rest on tuning this hyperparameter; they reflect \emph{what} the edge demonstrations are, not a search over how to weight them. Beyond this split, we examine the sensitivity to the two remaining numeric hyperparameters, the demonstration budget $K$ and the selection window $w$, in Appendix~\ref{app:kw}, where the advantage over both baselines stays stable across a wide range of each.

\begin{table}[tb!]
\centering
\caption{Allocation robustness. F1 (weighted F1 for the multi-class
DI) at each data-edge/model-edge split of the $K\!=\!5$ budget, per dataset; the
shaded cell is the best split and $\Delta$ is its gain over the fixed $2/3$ default. 
The best split beats $2/3$ by only $+0.010$
F1 on average, so \textsc{EdgeLM} is insensitive to the allocation. Rows are grouped by task (top to bottom: ED, EM, SM, DI, AD).}
\label{tab:allocation}
\footnotesize
\setlength{\tabcolsep}{3pt}
\begin{tabular}{@{}lcccccc@{\hspace{7pt}}c@{}}
\toprule
 & \multicolumn{6}{c}{Data edge / Model edge} & \\
\cmidrule(lr){2-7}
Dataset & 5/0 & 4/1 & 3/2 & 2/3 & 1/4 & 0/5 & $\Delta$\,2/3 \\
\midrule
Beers & 0.703 & 0.820 & 0.865 & \cellcolor{mygreen}0.913 & 0.907 & 0.886 & +0.000 \\
Fligh. & 0.563 & 0.644 & 0.660 & 0.669 & \cellcolor{mygreen}0.673 & 0.581 & +0.004 \\
Hosp. & 0.131 & 0.145 & 0.152 & 0.164 & \cellcolor{mygreen}0.170 & 0.163 & +0.006 \\
\midrule
A.-G. & 0.473 & \cellcolor{mygreen}0.493 & 0.481 & 0.472 & 0.436 & 0.444 & +0.021 \\
D.-A. & 0.737 & 0.820 & \cellcolor{mygreen}0.824 & 0.823 & 0.791 & 0.682 & +0.001 \\
W.-A. & 0.465 & 0.576 & 0.554 & 0.583 & 0.589 & \cellcolor{mygreen}0.590 & +0.007 \\
\midrule
ChEM. & 0.885 & 0.914 & \cellcolor{mygreen}0.923 & 0.906 & 0.904 & 0.829 & +0.017 \\
OpenD. & 0.928 & 0.917 & 0.946 & \cellcolor{mygreen}0.949 & 0.926 & 0.832 & +0.000 \\
TPC-D. & 0.916 & 0.918 & 0.939 & 0.970 & \cellcolor{mygreen}0.974 & 0.908 & +0.004 \\
\midrule
Balan. & 0.567 & 0.565 & \cellcolor{mygreen}0.608 & 0.565 & 0.563 & 0.558 & +0.043 \\
Diab. & 0.672 & 0.686 & \cellcolor{mygreen}0.687 & 0.655 & 0.550 & 0.520 & +0.032 \\
Splice & 0.517 & \cellcolor{mygreen}0.609 & 0.583 & 0.590 & 0.598 & 0.515 & +0.019 \\
\midrule
DoS & 0.854 & 0.934 & 0.954 & \cellcolor{mygreen}0.982 & 0.974 & 0.916 & +0.000 \\
Probe & 0.784 & 0.958 & 0.962 & \cellcolor{mygreen}0.977 & 0.934 & 0.831 & +0.000 \\
Rare & 0.816 & 0.949 & 0.960 & \cellcolor{mygreen}0.982 & 0.938 & 0.806 & +0.000 \\
\bottomrule
\end{tabular}
\vspace{-4mm}
\end{table}

\section{Conclusion}
\label{sec:conclusion}
We revisited demonstration selection for in-context table understanding and argued that its usual objective, similarity to the query, is poorly suited to instances near a decision boundary, where the nearest neighbors tend to share the query's likely label and reinforce it rather than expose the distinction the model must make. We instead proposed selecting edge demonstrations that lie close to a boundary, located from two complementary views, the labeled data and the deployed model's predictions. Casting five data wrangling tasks as one prediction problem, we evaluated on fifteen datasets and five LLMs, both open-source and proprietary. \textsc{EdgeLM} consistently outperforms the other selection strategies, and its two views prove complementary and robust to how the budget is divided. A demonstration's position relative to the decision boundary can therefore matter more than how close it is to the query, making boundary-aware retrieval a simple and general lever for in-context learning.

\section*{Limitations}
\textsc{EdgeLM} draws its demonstrations from a labeled pool and the frozen model's own predictions, so it assumes a modest amount of task supervision; where no labels exist, neither view of the boundary can be built. We consider this requirement light, since only a small pool is needed and no fine-tuning is performed. Our study also stays within tabular data wrangling under a single prediction abstraction; because the method depends only on retrieval over a labeled pool and the frozen model, with no task-specific component, extending it to streaming or online settings and to non-tabular structured data is a natural direction for future work.



\bibliography{ref, datasets}

\appendix

\label{sec:appendix}

\newtcblisting{promptbox}[1]{%
  listing only, breakable, enhanced,
  colback=gray!4, colframe=gray!55!black, coltitle=white,
  colbacktitle=gray!55!black, fonttitle=\bfseries\small, title={#1},
  boxrule=0.5pt, arc=2pt, left=4pt, right=4pt, top=3pt, bottom=3pt,
  listing options={basicstyle=\ttfamily\scriptsize, breaklines=true,
    breakatwhitespace=true, breakindent=0pt, columns=fullflexible,
    keepspaces=true, showstringspaces=false}%
}

\section{Prompt Templates}
\label{app:prompts}

Every task shares one few-shot scaffold: the task instruction, a \texttt{<solved examples>} block, and the query.
\textsc{EdgeLM} fills the block with data-edge (label-diverse) and model-edge demonstration (cases the frozen model itself answered wrong zero-shot, shown wrong$\rightarrow$right); \texttt{``..."} marks further demonstrations of the same component omitted for brevity. The examples are slices drawn from the real datasets.

\begin{promptbox}{Prompt Skeleton}
<task instruction ends by asking for a JSON answer>
Solved examples:
# data edge, label-diverse
<diverse demo, label A>       
Answer: {<gold JSON>}
<diverse demo, label B>
Answer: {<gold JSON>}
...
# model edge, wrong->right
<a case the model got wrong zero-shot>    
A tempting but WRONG answer: {<wrong JSON>}
Answer: {<gold JSON>}
...
Now solve:
<query content>
\end{promptbox}

\begin{promptbox}{Error Detection}
Task: decide whether the TARGET cell value is a data error (typo, format violation, or value inconsistent with the rest of its column and row). Answer in JSON: {"error": "yes"} or {"error": "no"}.
Solved examples:
Column: city
Target cell value: 'birmingham'
Row: index: 1 | provider_number: 10018 | name: callahan eye foundation hospital | address_1: 1720 university blvd | city: birmingham | state: al | zip: 35233 | ...
Answer: {"error": "no"}
...
Column: city
Target cell value: 'birminghxm'
Row: index: 5 | provider_number: 10018 | name: callahan eye foundation hospital | address_1: 1720 university blvd | city: birminghxm | state: al | zip: 35233 | ...
Answer: {"error": "yes"}
...
Column: city
Target cell value: 'boaz'
Row: index: 57 | provider_number: 10005 | name: marshall medical center south | address_1: 2505 u s highway 431 north | city: boaz | state: al | zip: 35957 | ...
A tempting but WRONG answer: {"error": "yes"}
Answer: {"error": "no"}
...
Now solve:
Column: city
Target cell value: 'fxorence'
Row: index: 80 | provider_number: 10006 | name: eliza coffee memorial hospital | address_1: 205 marengo street | city: fxorence | state: al | zip: 35631 | ...
\end{promptbox}

\begin{promptbox}{Entity Matching}
Task: decide whether the two records refer to the same real-world entity. Answer in JSON: {"label": "match"} or {"label": "non-match"}.
Solved examples:
Entity A: title: emergent semantics and the multimedia semantic web | authors: william i. grosky , farshad fotouhi , d. v. sreenath | venue: sigmod record | year: 2002
Entity B: title: emergent semantics and the multimedia semantic web | authors: w. i. grosky , d. v. sreenath , f. fotouhi | venue: acm sigmod record | year: 2002
Answer: {"label": "match"}
...
Entity A: title: the aditi deductive database system | authors: kotagiri ramamohanarao , david b. kemp | venue: vldb j. | year: 1994
Entity B: title: space optimization in deductive databases | authors: divesh srivastava , s. sudarshan | venue: acm transactions on database systems ( tods ) | year: 1995
Answer: {"label": "non-match"}
...
Entity A: title: web caching for database applications with oracle web cache | authors: jordan parker , jesse anton , zheng zeng | venue: sigmod conference | year: 2002
Entity B: title: form-based proxy caching for database-backed web sites | authors: qiong luo , jeffrey f. naughton | venue: very large data bases | year: 2001
A tempting but WRONG answer: {"label": "match"}
Answer: {"label": "non-match"}
...
Now solve:
Entity A: title: mining association rules between sets of items in large databases | authors: r. agrawal , t. imielinski , a. swami | venue: sigmod conference | year: 1993
Entity B: title: mining association rules between sets of items in large databases | authors: rakesh agrawal , tomasz imielinski , arun swami | venue: international conference on management of data | year: 1993
\end{promptbox}

\begin{promptbox}{Schema Matching}
Task: decide whether Column A and Column B refer to the same attribute (would map to each other in schema matching). Answer in JSON: {"label": "match"} or {"label": "non-match"}.
Solved examples:
Column A: tid
Values: 12052 | 22226 | 104729 | 80001 | 50185
Column B: assays_td
Values: 50597 | 22229 | 104729 | 80001
Answer: {"label": "match"}
...
Column A: assay_id
Values: 1 | 2 | 3 | 4 | 5
Column B: assays_cnfdnc_scr
Values: 1 | 0
Answer: {"label": "non-match"}
...
Column A: doc_id
Values: 11087 | 684 | 15453 | 17841 | 17430
Column B: assay_id
Values: 1 | 2 | 3 | 4 | 5
A tempting but WRONG answer: {"label": "match"}
Answer: {"label": "non-match"}
...
Now solve:
Column A: assay_organism
Values: Homo sapiens | Mus musculus | Rattus norvegicus
Column B: assays_organism
Values: Homo sapiens | Sus scrofa | Bos taurus
\end{promptbox}

\begin{promptbox}{Data Imputation}
Task: impute the missing value (marked <MISSING>) using the labeled reference rows. Answer in JSON: {"value": "<answer>"}.
Solved examples:
preg: 1.0 | plas: 97.0 | pres: 66.0 | skin: 15.0 | insu: 140.0 | mass: 23.2 | pedi: 0.487 | age: 22.0 -> class: <MISSING>
Answer: {"value": "tested_negative"}
...
preg: 6.0 | plas: 148.0 | pres: 72.0 | skin: 35.0 | insu: 0.0 | mass: 33.6 | pedi: 0.627 | age: 50.0 -> class: <MISSING>
Answer: {"value": "tested_positive"}
...
preg: 3.0 | plas: 180.0 | pres: 64.0 | skin: 25.0 | insu: 70.0 | mass: 34.0 | pedi: 0.271 | age: 26.0 -> class: <MISSING>
A tempting but WRONG answer: {"value": "tested_positive"}
Answer: {"value": "tested_negative"}
...
Now solve:
preg: 7.0 | plas: 159.0 | pres: 64.0 | skin: 0.0 | insu: 0.0 | mass: 27.4 | pedi: 0.294 | age: 40.0 -> class: <MISSING>
The value must be one of: tested_negative, tested_positive
\end{promptbox}

\begin{promptbox}{Anomaly Detection}
Task: decide whether the data record is a normal record or an anomaly (an outlier inconsistent with typical records). Answer in JSON: {"label": "normal"} or {"label": "anomaly"}.
Solved examples:
duration: 0 | protocol_type: tcp | service: http | flag: SF | src_bytes: 301 | dst_bytes: 351 | ...
Answer: {"label": "normal"}
...
duration: 0 | protocol_type: tcp | service: private | flag: S0 | src_bytes: 0 | dst_bytes: 0 | ...
Answer: {"label": "anomaly"}
...
duration: 0 | protocol_type: icmp | service: ecr_i | flag: SF | src_bytes: 520 | dst_bytes: 0 | ...
A tempting but WRONG answer: {"label": "normal"}
Answer: {"label": "anomaly"}
...
Now solve:
duration: 0 | protocol_type: tcp | service: http | flag: SF | src_bytes: 317 | dst_bytes: 820 | ...
\end{promptbox}

\section{Per-Dataset Results}
\label{app:perdataset}

Table~\ref{app:tab:perdataset} gives the per-dataset numbers behind the per-task averages of Table~\ref{tab:main} for Llama-3.1-8B. \textsc{EdgeLM} attains the best or tied-best F1 on 13 of the 15 datasets, and its margins are largest exactly where the similarity baseline is weakest. For example, on the three anomaly-detection datasets it lifts F1 from $0.78$--$0.85$ to $0.98$ and accuracy from $0.89$--$0.93$ to $0.99$. These tasks have a genuine local decision boundary that the contrast pair and the corrected mistakes make explicit, so the extra structure translates directly into both metrics.

On a few datasets Edge ties the similarity baseline on F1 and accuracy alike (Amazon-Google at $0.473$/$0.853$, BalanceScale at $0.567$/$0.579$, Splice at $0.517$/$0.530$). These are lookup-like instances where the answer already lies in the nearest neighbor and there is no local boundary to sharpen, so Edge matches it by construction instead of risking a regression.

Two datasets favor a baseline, and on both it is the label-diverse baseline that overtakes Edge. On DBLP-ACM, the frozen model already matches these canonical bibliographic records almost perfectly without demonstrations ($0.855$ F1 / $0.946$ accuracy). The diverse baseline nudges F1 up to $0.894$ ($0.963$ accuracy), while Edge's boundary and mistake demonstrations add surface-form patterns that slightly perturb an already-correct default ($0.824$ F1 / $0.941$ accuracy) rather than helping. On Hospital the diverse baseline leads outright ($0.221$ F1 / $0.798$ accuracy), ahead of other techniques. Hospital is extremely imbalanced (about $2.5\%$ of cells are errors) and its errors are subtle single-character corruptions; because errors are so rare, a query's nearest neighbors are almost always clean, so both similarity and Edge's locally label-diverse contrast surface mostly clean rows and push the model toward the majority ``clean'' label, whereas balancing the label space across the whole pool, exactly what the diverse baseline does, is the one strategy that reliably surfaces error examples.

\begin{table*}[t!]
\centering
\footnotesize
\caption{Per-dataset results for Llama-3.1-8B. Each cell is F1 / weighted F1 (Accuracy in parentheses); \textbf{bold} marks the best
F1 per row.}
\label{app:tab:perdataset}
\setlength{\tabcolsep}{3pt}
\resizebox{\textwidth}{!}{\footnotesize
\begin{tabular}{@{}llccccc@{}}
\toprule
Task & Dataset & Zero-shot & Random & Diversity & Similarity & Edge \\
\midrule
\multirow{3}{*}{ED} & Beers~\cite{mahdavi2019raha} & 0.379 (0.759) & 0.448 (0.873) & 0.542 (0.863) & 0.726 (0.909) & \textbf{0.907 (0.973)} \\
 & Flights~\cite{ds:flights} & 0.373 (0.696) & 0.390 (0.805) & 0.472 (0.807) & 0.565 (0.787) & \textbf{0.665 (0.858)} \\
 & Hospital~\cite{ds:hospital} & 0.104 (0.553) & 0.197 (0.769) & \textbf{0.221 (0.798)} & 0.132 (0.617) & 0.171 (0.717) \\
\midrule
\multirow{3}{*}{EM} & Amazon-Google~\cite{kopcke2010evaluation} & 0.277 (0.602) & 0.353 (0.732) & 0.302 (0.649) & \textbf{0.473 (0.853)} & \textbf{0.473 (0.853)} \\
 & DBLP-ACM~\cite{kopcke2010evaluation} & 0.855 (0.946) & 0.765 (0.902) & \textbf{0.894 (0.963)} & 0.737 (0.891) & 0.824 (0.941) \\
 & Walmart-Amazon~\cite{ds:magellan} & 0.237 (0.455) & 0.272 (0.553) & 0.253 (0.516) & 0.465 (0.821) & \textbf{0.583 (0.905)} \\
\midrule
\multirow{3}{*}{SM} & ChEMBL-Join~\cite{koutras2021valentine} & 0.723 (0.842) & 0.729 (0.825) & 0.732 (0.839) & 0.885 (0.919) & \textbf{0.923 (0.947)} \\
 & OpenData-Join~\cite{koutras2021valentine} & 0.812 (0.882) & 0.821 (0.872) & 0.812 (0.865) & 0.928 (0.951) & \textbf{0.946 (0.964)} \\
 & TPC-DI-Join~\cite{koutras2021valentine} & 0.833 (0.902) & 0.850 (0.901) & 0.852 (0.903) & 0.916 (0.938) & \textbf{0.937 (0.955)} \\
\midrule
\multirow{3}{*}{DI} & BalanceScale~\cite{ds:balancescale} & 0.006 (0.056) & 0.494 (0.528) & 0.499 (0.493) & \textbf{0.567 (0.579)} & \textbf{0.567 (0.579)} \\
 & DiabetesPima~\cite{ds:pima} & 0.176 (0.344) & 0.616 (0.628) & 0.649 (0.652) & 0.672 (0.673) & \textbf{0.687 (0.701)} \\
 & Splice~\cite{ds:splice} & 0.087 (0.231) & 0.336 (0.328) & 0.257 (0.301) & \textbf{0.517 (0.530)} & \textbf{0.517 (0.530)} \\
\midrule
\multirow{3}{*}{AD} & KDD-DoS~\cite{ds:kddcup99} & 0.203 (0.703) & 0.641 (0.790) & 0.618 (0.726) & 0.852 (0.928) & \textbf{0.982 (0.992)} \\
 & KDD-Probe~\cite{ds:kddcup99} & 0.600 (0.815) & 0.605 (0.760) & 0.616 (0.731) & 0.784 (0.888) & \textbf{0.977 (0.988)} \\
 & KDD-RareAttack~\cite{ds:kddcup99} & 0.236 (0.706) & 0.405 (0.677) & 0.492 (0.633) & 0.794 (0.902) & \textbf{0.981 (0.991)} \\
\bottomrule
\end{tabular}}
\vspace{-4mm}
\end{table*}

\section{Computational Cost}
\label{app:cost}
Relative to plain similarity retrieval, \textsc{EdgeLM} adds computation in only two places: a one-time construction of the model-error pool and a slightly heavier per-query retrieval. We separate the cost into language-model calls and retrieval, a set of vector operations that we measure directly on the demonstration pool.

In terms of language-model calls, the selection strategies are identical at inference. Each query issues a single $K$-shot call, and \textsc{EdgeLM} places the same number of demonstrations in the prompt as similarity, random, or diverse selection, so the dominant cost is
unchanged. The only additional calls \textsc{EdgeLM} incurs are a one-time construction of the model-error pool, a zero-shot pass over the demonstration pool, that is $|\mathrm{pool}|$ calls. These are made once, reused for every subsequent query, and grow linearly with the pool, so amortised over a deployment they add nothing to the per-query call count; similarity, random, and diverse selection make no such calls.

The retrieval step differs only in vector operations, and only mildly. Random selection draws $K$ indices from the pool with no similarity computation and no ranking, so its cost is essentially zero. Diverse selection needs no similarity computation and no sort; it partitions the pool by label ($O(N)$) and draws a label-balanced sample, so its cost is of the same negligible order as random selection. Similarity retrieval computes the query-to-pool similarities ($O(Nd)$ for a pool of $N$ examples in $d$ dimensions) and sorts them to take the $K$ nearest ($O(N\log N)$). \textsc{EdgeLM} reuses that same ranking, then (i) groups the top-$w$ neighbors by label and selects round-robin across the groups to obtain the label-diverse demonstrations ($O(w)$), and (ii) performs a second nearest-neighbor search over the smaller error pool, adding one more similarity-and-sort of $O(N_e d + N_e\log N_e)$ with $N_e\le N$. Both methods are thus dominated by the same $O(Nd)$ similarity computation and $O(N\log N)$ sort; edge adds only a second, smaller sort and a constant-time grouping.

Table~\ref{app:tab:cost} reports the measured per-query retrieval latency as the pool grows. The overhead is a small constant factor and, like similarity retrieval, scales linearly with the pool; at a pool of $2{,}000$ edge retrieval takes $0.13$\,ms against $0.06$\,ms for similarity, and even at a pool of $8{,}000$ it stays under half a millisecond. Since every strategy issues one $K$-shot call per query, this
sub-millisecond difference in vector operations does not affect the number of model calls, which is what governs the end-to-end cost.

A final cost is optional. The split of the budget between the two components, the data edge and the model edge, is itself a hyperparameter, but one we fix rather than tune. Section~\ref{sec:robustness} shows the method is insensitive to it. Should one nonetheless wish to tune it per dataset or model, the cost is a one-time, offline pass. We set aside a small stratified fraction of the demonstration pool as a held-out slice and, for each of the $K$ candidate allocations, score that slice under the gated selection, keeping the one with the highest held-out accuracy. This reuses the already-mined error pool, so it requires no further mining; its only cost is $O(K|H|)$ language-model calls for a held-out slice of size $|H|$, a small capped fraction of the pool, again incurred once and amortised over all queries.

Overall, the per-query language-model cost is identical across selection strategies, the added retrieval work is a second sort over a smaller pool and stays sub-millisecond, and the only extra language-model work is a one-time, pool-linear mining pass. The overhead of \textsc{EdgeLM} is therefore modest and confined to a one-time setup. 

\begin{table}[t]
\centering
\small
\caption{Per-query retrieval latency (milliseconds, averaged over the five tasks) as the demonstration
pool grows. Edge reuses similarity's ranking and adds a second sort over the smaller error
pool; both scale linearly with the pool. Random selection needs no similarity computation or
sort and is omitted (its cost is $\approx 0$).}
\label{app:tab:cost}
\begin{tabular}{@{}rrrr@{}}
\toprule
Pool size & Similarity (ms) & Edge (ms) & Overhead (ms) \\
\midrule
250   & 0.010 & 0.029 & +0.019 \\
500   & 0.018 & 0.040 & +0.022 \\
1{,}000  & 0.032 & 0.067 & +0.035 \\
2{,}000  & 0.060 & 0.126 & +0.066 \\
4{,}000  & 0.093 & 0.233 & +0.141 \\
8{,}000  & 0.192 & 0.463 & +0.271 \\
\bottomrule
\end{tabular}
\vspace{-4mm}
\end{table}

\section{Sensitivity to the Demonstration Budget and the Selection Window}
\label{app:kw}

We sweep $K\in\{3,5,8,10\}$ and $w\in\{5,\allowbreak 10,\allowbreak 20,\allowbreak 50,\allowbreak 100\}$ on all fifteen datasets with Llama-3.1-8B. Because the grid spans $20$ configurations over $15$ datasets, and each cell re-scores the full test set, this experiment is considerably larger than the main comparison; we therefore run it with a single seed.

\begin{figure*}[t]
\centering
\begin{subfigure}{\textwidth}
    \centering
    \includegraphics[width=\textwidth]{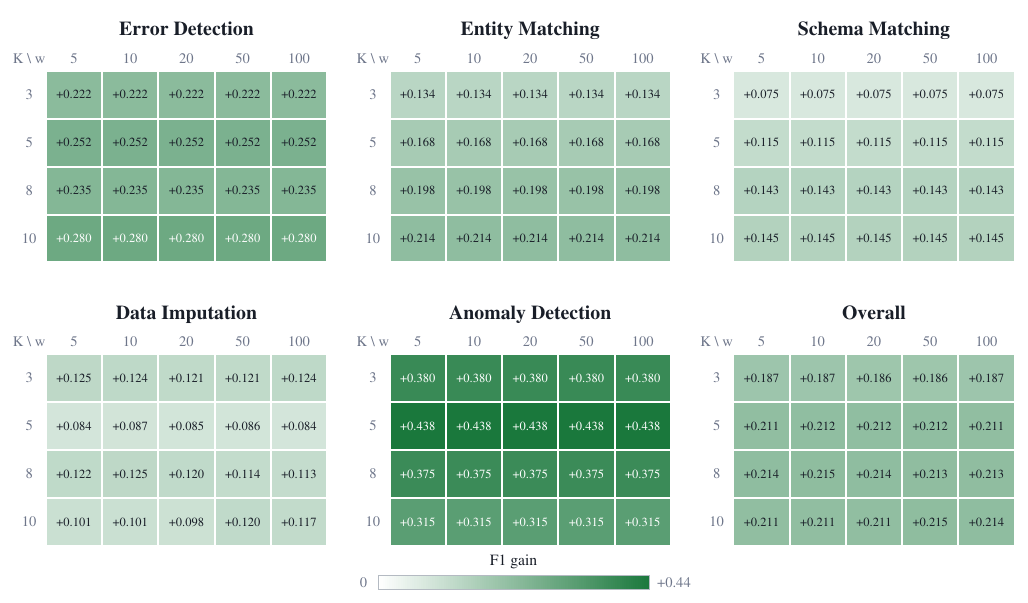}
    \caption{Gain over \emph{random} selection.}
    \label{app:fig:kw-rand}
\end{subfigure}

\vspace{8pt}

\begin{subfigure}{\textwidth}
    \centering
    \includegraphics[width=\textwidth]{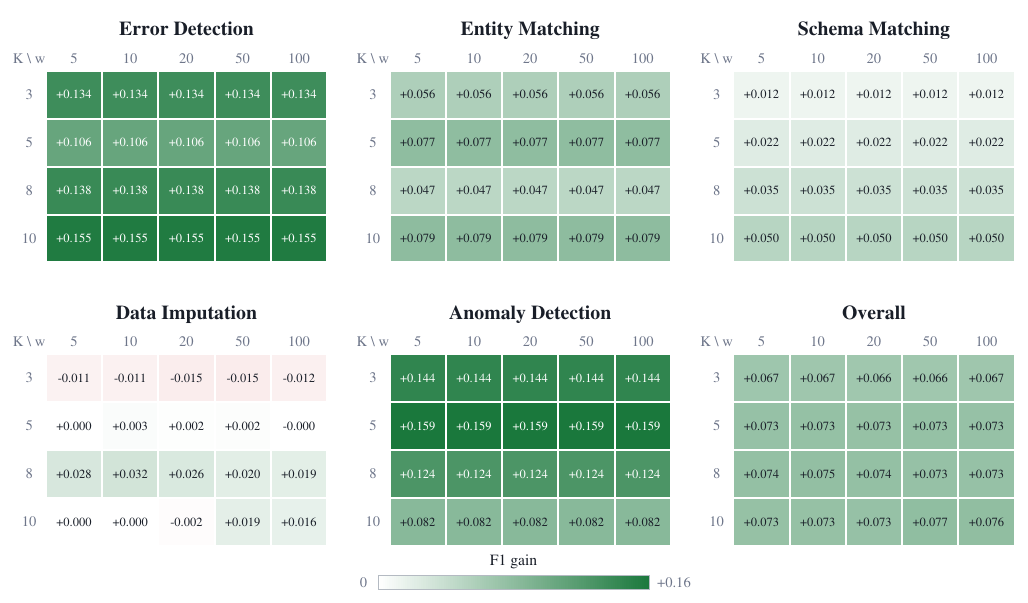}
    \caption{Gain over \emph{similarity} selection.}
    \label{app:fig:kw-sim}
\end{subfigure}
\caption{F1 gain of \textsc{EdgeLM} over random and similarity selection across the demonstration budget $K$ (rows) and the selection window $w$ (columns), for Llama-3.1-8B on all fifteen datasets.}
\label{app:fig:kw}
\end{figure*}

Figure~\ref{app:fig:kw} reports the F1 gain of edge selection over random and similarity baselines. Against random selection (Figure~\ref{app:fig:kw-rand}) all cells are positive and the task-averaged gain, shown in the overall panel, stays between $+0.186$ and $+0.215$ across the entire grid. Against the stronger similarity baseline (Figure~\ref{app:fig:kw-sim}) the task-averaged gain stays between $+0.066$ and $+0.077$, a spread of about $0.01$ over a grid that varies $K$ by more than a factor of three and $w$ by a factor of twenty. The per-task picture is equally stable. Over similarity, error detection ranges from $+0.106$ to $+0.155$ and anomaly detection from $+0.082$ to $+0.159$, while entity matching stays within $+0.047$ to $+0.079$ and schema matching within $+0.012$ to $+0.050$.

Two patterns are worth noting. First, the gains are essentially unaffected by $w$. Once the gate has identified a boundary among its nearest neighbors, the label-diverse demonstrations already lie inside that neighborhood, so widening $w$ selects much the same examples. Data imputation, which is multi-class, is the only task that shifts with $w$, and only slightly. Second, the only non-positive cells are in data imputation (F1 gain from $-0.015$ to $+0.032$). Its two lookup-style datasets (BalanceScale and Splice) take an all-diverse allocation at the smaller budgets, so the diverse set coincides with the $K$ nearest neighbors and edge reduces to similarity selection; as the budget grows the mistake component re-enters and the gain turns positive.

The method is therefore insensitive to both hyperparameters. No setting of $(K,w)$ in this range alters the conclusion that edge selection beats both baselines, and the task-averaged advantage varies by less than one F1 point. This is what lets the main experiments fix a single small budget, $K\!=\!5$, without tuning it per task or per model.

\end{document}